\documentclass[journal]{IEEEtran}
\usepackage{blindtext, graphicx}
\usepackage{amsmath}
\usepackage{fancyhdr}

\usepackage[backend=bibtex,style=ieee,sorting=ynt]{biblatex}
\begin{document}
\title{Semi-Supervised Radio Signal Identification}

\author{
\IEEEauthorblockN{Timothy J. O'Shea\IEEEauthorrefmark{1},
    Nathan West\IEEEauthorrefmark{2}, 
    Matthew Vondal\IEEEauthorrefmark{1} and
    T. Charles Clancy\IEEEauthorrefmark{1}}
\IEEEauthorblockA{\IEEEauthorrefmark{1}Bradley Department of Electrical and Computer Engineering, Virginia Tech
    \\\{oshea, mvondal, tcc\}@vt.edu}
\IEEEauthorblockA{\IEEEauthorrefmark{2}School of Electrical and Computer Engineering, Oklahoma State University
    \\\{nathan.west\}@okstate.edu}
}

\maketitle

\begin{abstract}
Radio emitter recognition in dense multi-user environments is an important tool for optimizing spectrum utilization, identifying and minimizing interference, and enforcing spectrum policy.  Radio data is readily available and easy to obtain from an antenna, but labeled and curated data is often scarce making supervised learning strategies difficult and time consuming in practice.  We demonstrate that semi-supervised learning techniques can be used to scale learning beyond supervised datasets, allowing for discerning and recalling new radio signals by using sparse signal representations based on both unsupervised and supervised methods for nonlinear feature learning and clustering methods.
\end{abstract}


\section{Introduction}

Radio signal recognition in dense and complex multi-user spectrum environments is an important tool for optimizing spectrum utilization, identifying and minimizing interference, enforcing spectrum policy, and implementing effective radio sensing and coordination systems.  Classical approaches to the problem focus on energy detection and the use of expert features and decision criteria to identify and categorize specific modulation types \cite{hsue1989automatic} \cite{gardner1988cyclic}.  These approaches rely on prior knowledge of signal properties, features, and decision statistics to separate known modulations and are typically derived under simplified analytic hardware, propagation, radio environment models.

We recently demonstrated the viability of naive feature learning for supervised radio classification systems \cite{convmodrec} which allows for joint feature and classifier learning given labeled datasets and examples.  In this case we were able to outperform traditional expert decision statistic based classification in sensitivity and accuracy by a significant margin.

This was a powerful result, providing significant performance improvements against current day solutions, but it still relied entirely on supervised learning and well curated training data.  In the real world, and especially in the radio domain, we are faced with vast amounts of unlabeled example data available to our sensor and incomplete knowledge of class labels comprising ground truth.

To address this problem we investigate alternative strategies for radio identification learning which rely less heavily on labeled training data and are capable of making sense of radio signals with either no or less labeled examples, potentially drastically reducing the burden of data curation on such a machine learning system for developers and maintainers, and allowing systems to recognize new signals and scale to to understand new environments over time.

\section{Background}

In semi-supervised learning \cite{semisup1} \cite{semisup2} \cite{semisup3} we seek to to separate and identify new classes without explicit class labels on examples of these classes allowing us to specifically learn features to separate them.  

To approach this problem, we consider performing dimensionality reduction on signal examples into a smooth smaller space where we can perform signal clustering.  Given an appropriate dimensionality reduction, we seek a space where signals of the same or similar type have a low distance separating them while signals of differing types are separated by larger distances.  Ideally in such a space, examples of the same or similar types form discrete and separable clusters that are readily discernible from each other. 

Classification of signal types in such a space then becomes a problem of identifying clusters, associating a label with each cluster (rather than each example, a much less labor intensive task), and one that allows for the recognition, classification, and historical annotation of new classes and new class examples even without knowledge of the label.

\section{Learning Sparse Representations}

\begin{figure}[h]
    \centering
    \includegraphics[width=0.45\textwidth]{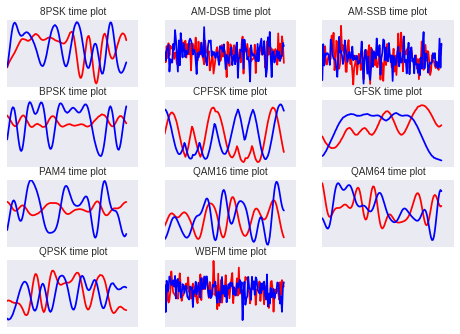}
    \caption{Examples from the dataset}
    \label{figure:dataset}
\end{figure}

We focus on learning sparse representations of raw sampled radio signal time series examples using two forms of convolutional neural networks.  We leverage the RadioML16.04 \cite{radioml} labeled radio modulation dataset over 11 modulations which includes effects of white noise, oscillator drift, sample clock drift, and fading.  Examples from this data set are shown in figure \ref{figure:dataset}.

\subsection{Purely Unsupervised Sparse Representation}

In the case where no class labels are used, we take a purely unsupervised approach to learning a sparse representation of the dataset.  This can be done by applying dependence based dimensionality reduction techniques such as Principal Component Analysis (PCA) or Independent Component Analysis (ICA), however instead we leverage a non-linear dimensionality reduction given by reconstruction of the input signal through a sparse representation over a set of convolutional basis functions \cite{convnn} learned within an autoencoder neural network \cite{masci2011stacked}.  Autoencoders are unsupervised learning constructs in which the optimization goal of the neural network is to minimize reconstruction error at the output to match the input, through some intermediate representation of a more constrained dimension, typically a mean squared error (MSE) loss function is used and a form of stochastic gradient descent is used as a solver by back-propagating gradients from this loss term to find best network parameters which approximate those in equation \ref{equation:eq1}.  We optimize using RMSProp \cite{rmsprop} and Adam \cite{adam} gradient descent solvers, which both obtain relatively similar results.

\begin{equation}
    \underset{\theta}{argmin} \left( \sum  \left(X-f\left(X,\theta \right) \right)^2 \right)
\end{equation}\label{equation:eq1}

\begin{figure}[h]
    \centering
    \includegraphics[width=0.5\textwidth]{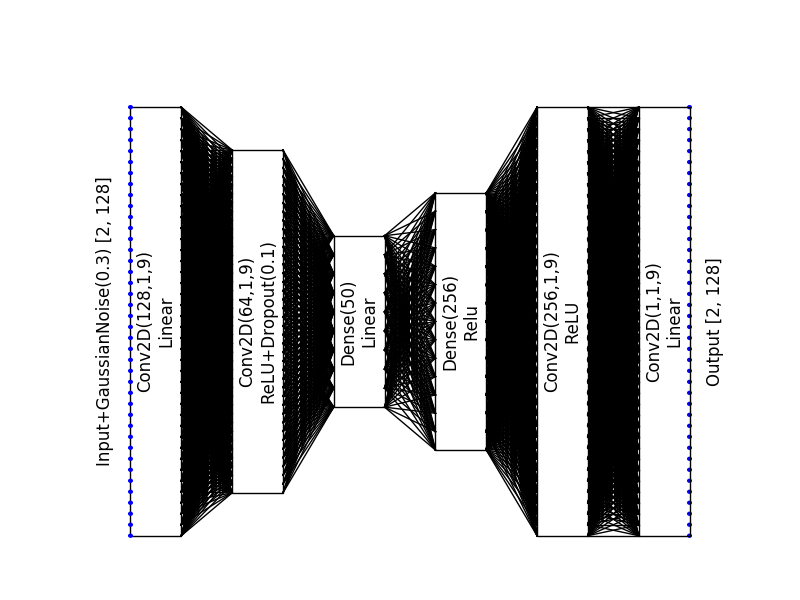}
    \caption{Convolutional AutoEncoder Structure}
    \label{figure:convae_arch}
\end{figure}

By constraining intermediate layer widths within the network from which the original full-dimension examples can be reconstructed with low error, we obtain a non-linear dimensionality reduction by extracting the intermediate sparse encoding to use for clustering.  In this case, modulations using similar basis functions and pulse shapes can be represented by similarly shaped convolutional filters and intermediate feature maps within the encoder and decoder, leading them to exist in similar regions of this compressed space.  The convolutional layers in the autoencoder are well suited for radio time series signal representation due to their shift invariant representation properties in time and the constrained parameter search space vs their fully-connected layer equivalents.  We use both dropout \cite{srivastava2014dropout} and input noise \cite{vincent2008extracting} as regularization techniques to help improve generalization of our learned representation on the dataset.  The architecture used for our convolutional autoencoder is shown in figure \ref{figure:convae_arch}

\begin{figure}[h]
    \centering
    \includegraphics[width=0.5\textwidth]{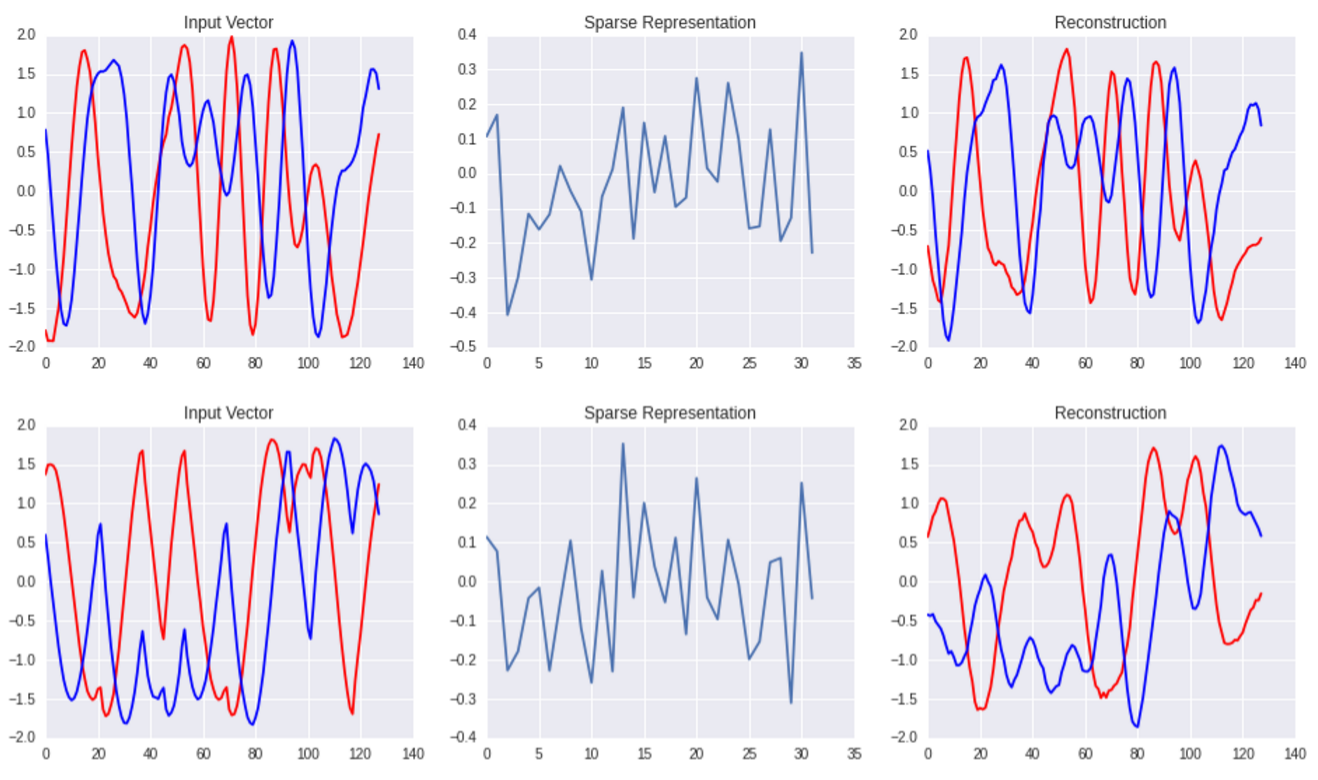}
    \caption{Examples from the ConvAE}
    \label{figure:convAE}
\end{figure}

During training of the autoencoder, we minimize reconstruction mean square error (MSE), but since our primary goal is obtaining a good sparse representation for clustering, we significantly constrain the hidden layer dimension to a point where our reconstruction makes some visible simplifying assumptions, favoring decreased hidden layer dimension over optimal reconstruction error.  Figure \ref{figure:convAE} shows for two training examples what the 2x128 input vector looks like, what the 1x30 sparse representation looks like, and what the 2x128 output reconstruction looks like.  This gives some intuition as to the learned representational capacity of this network.

\subsection{Supervised Bootstrapping of Sparse Representation}

\begin{figure}[h]
    \centering
    \includegraphics[width=0.5\textwidth]{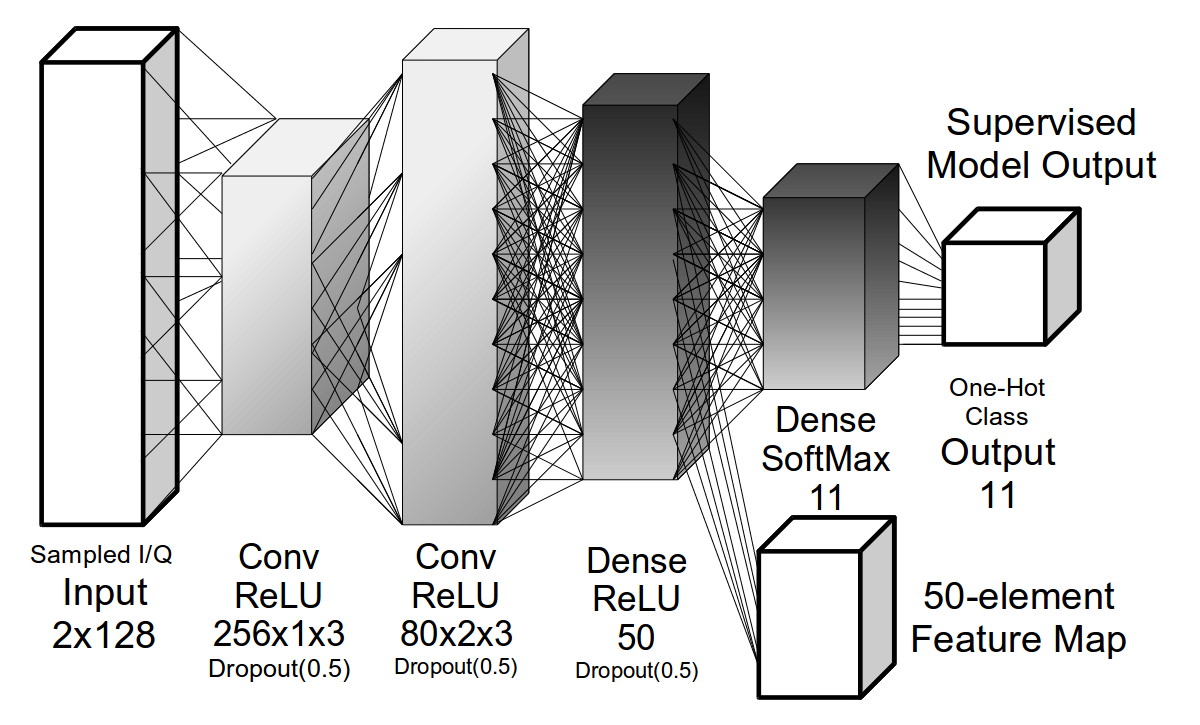}
    \caption{Architecture of the Supervised Bootstrap Network}
    \label{figure:supervised_arch}
\end{figure}

In the case that we do have some expertly labeled data, we can also generate a sparse representation space using discriminative features learned during supervised training.  In prior work \cite{convmodrec} we trained a convolutional neural network in a purely supervised way to to label examples, but here we leverage this trained network and discard the final softmax layer to keep only high level learned feature maps as sparse representations.  The networks used in supervised training and extraction of sparse representation using the learned feature-maps are shown in figure \ref{figure:supervised_arch}

Features formed in this way leverage and distill available expertly curated labels, but in many cases they also generalize and provide a capacity to separate additional classes in feature space which do not have class labels.  We therefore treat this as a bootstrap method for sparse feature learning in a supervised way and generalize to semi-supervised recognition over much larger data-sets containing additional unlabeled training data.

\section{Visualizing Signal-Type Embeddings}

To visualize these learned representations and their class separability, we perform t-distributed Stochastic Neighbor Embedding (t-SNE) \cite{tsne} on the compressed representations to display the examples on a 2 dimensional manifold where we can get some idea of the clustering and distances present in the underlying dataset.

\begin{figure}[ht!]
    \centering
    \includegraphics[width=0.5\textwidth]{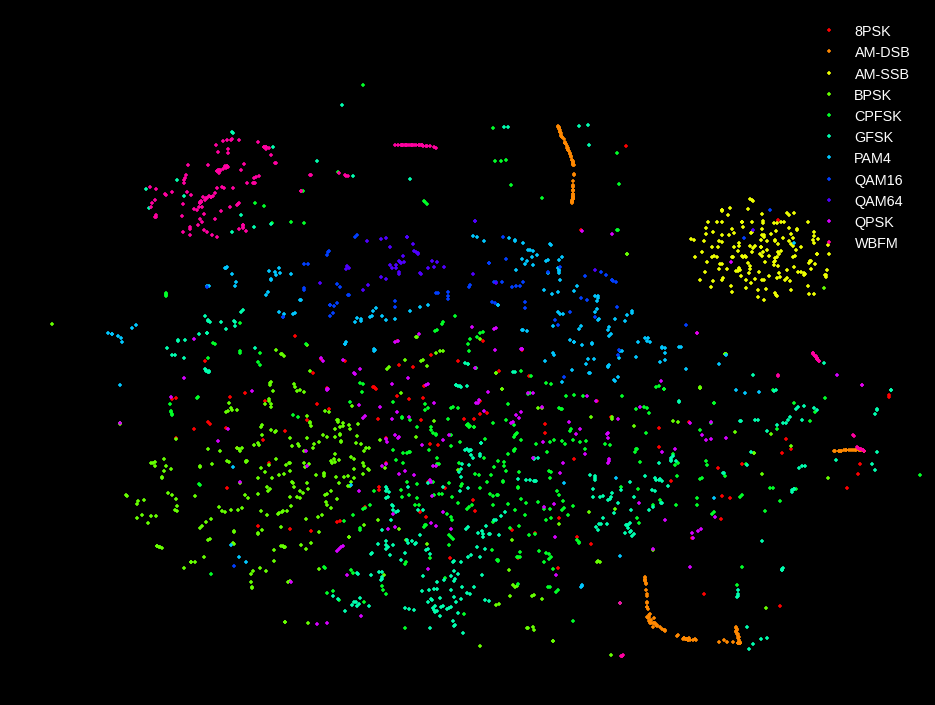}
    \caption{Modulation Embedding from ConvAE1 Features}
    \label{figure:ss1}
\end{figure}

\begin{figure}[ht!]
    \centering
    \includegraphics[width=0.5\textwidth]{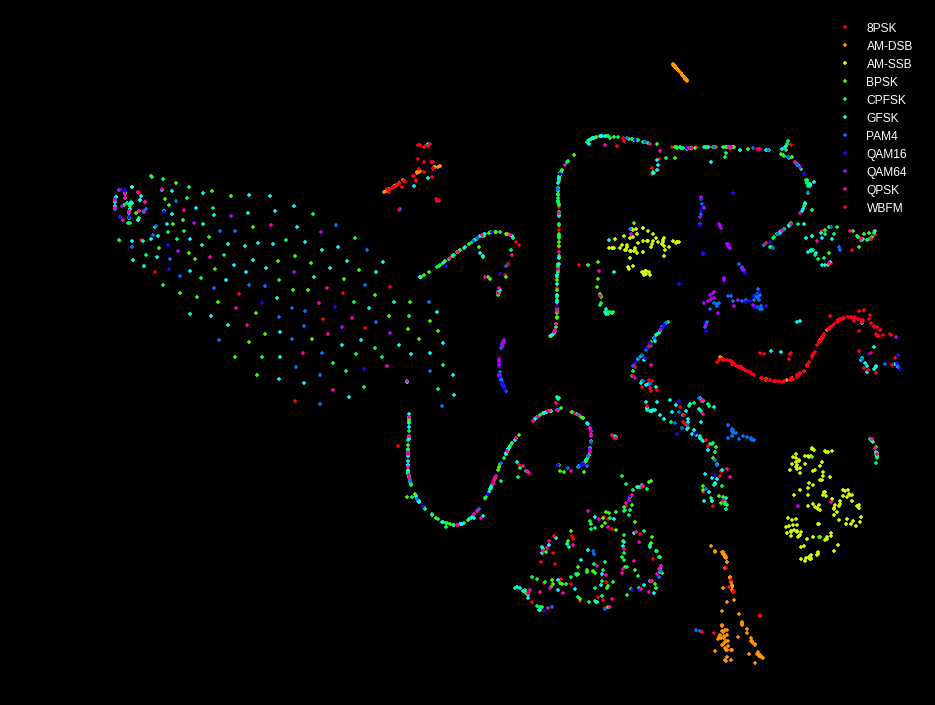}
    \caption{Modulation Embedding from ConvAE2 Features}
    \label{figure:ss3}
\end{figure}

We first look at the clustering and separability of classes from the t-SNE embedding of classes in the auto-encoder's representational feature space, shown in \ref{figure:ss1} and \ref{figure:ss3}.   In this case, we see that several of the classes such as WBFM, AM-DSB, AM-SSB, and QPSK (in ConvAE1) have formed distinct and mostly-separable clusters while other classes demonstrate significantly more mixing and would be difficult to separate through a clustering approach.  Still this is not bad considering the features were never trained to be discriminators yet we have obtained some level of class separability.  We believe a hierarchical approach to the separability of remaining classes, or some amount of iteration between unsupervised cluster learning and discriminative feature map learning on separable clusters may help, but is beyond the scope of this work.

\begin{figure}[h]
    \centering
    \includegraphics[width=0.5\textwidth]{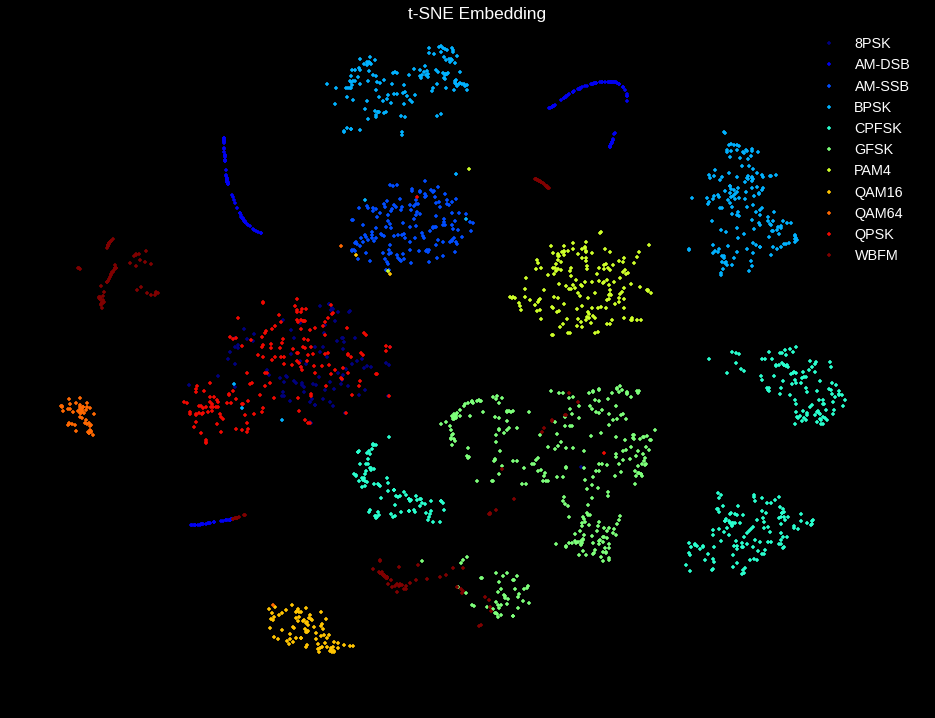}
    \caption{Modulation Embedding from Bootstrap Features}
    \label{figure:ss2}
\end{figure}

Second we look at the clustering of the t-SNE embeddings of the bootstrapped discriminative feature representation, shown in figure \ref{figure:ss2}.  In this case we obtain distinct and almost completely separable clustering for almost every modulation in the dataset.   

Of course here we have guilty class-discriminative knowledge used while forming this feature map representation (during supervised training), but ideally these discriminative features will continue to generalize over a large number of classes and help discriminate additional unknown signals as well.

\section{Learning Generalization}

To test this theory that learned discriminative features may generalize and help discriminate new unknown modulations in a semi-supervised way using the bootstrap approach, we repeat our prior approach but this time we train the supervised classifier on 9 out of the 11 modulations.  We hold out BPSK and 16QAM modulations, presenting no information about these classes or their examples during supervised training.  We then take examples from all 11 classes and transform them into the compressed feature-map space and visualize the 2-dimensional embedding using t-SNE shown below in figure \ref{figure:ss5}

\begin{figure}[h]
    \centering
    \includegraphics[width=0.5\textwidth]{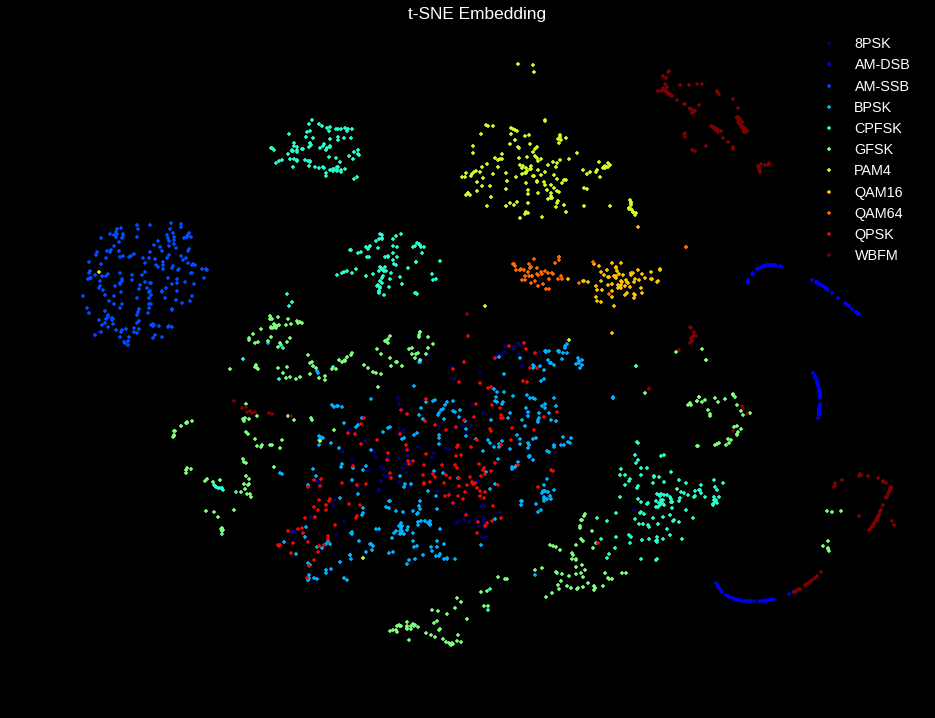}
    \caption{Bootstap Modulation Embedding (holdout BPSK and 16QAM) }
    \label{figure:ss5}
\end{figure}

Closely inspecting these results we can see that BPSK has been unfortunately quite heavily mixed with both QPSK and 8PSK modulations (which it is a subset of).   However, QAM16, a previously unseen class to these features, has been tightly clustered in the vicinity of QAM64 in a fairly well defined and separable region in the embedding space.   

These results, while very preliminary and qualitative, do support the fact that bootstrap feature-map representations do carry significant capacity for generalization to identifying, clustering, and discerning new unknown or previously unseen modulation types, but they do not guarantee clean separability in all cases.  We hope that as the number of classes and features scales, this generalization will improve.  One of the challenges for moving this field of study forward will be to identify and quantify measures for the ability of features to generalize well, and seek to focus on improving this metric.

\section{Class Clustering}

\begin{figure}[h]
    \centering
    \includegraphics[width=0.5\textwidth]{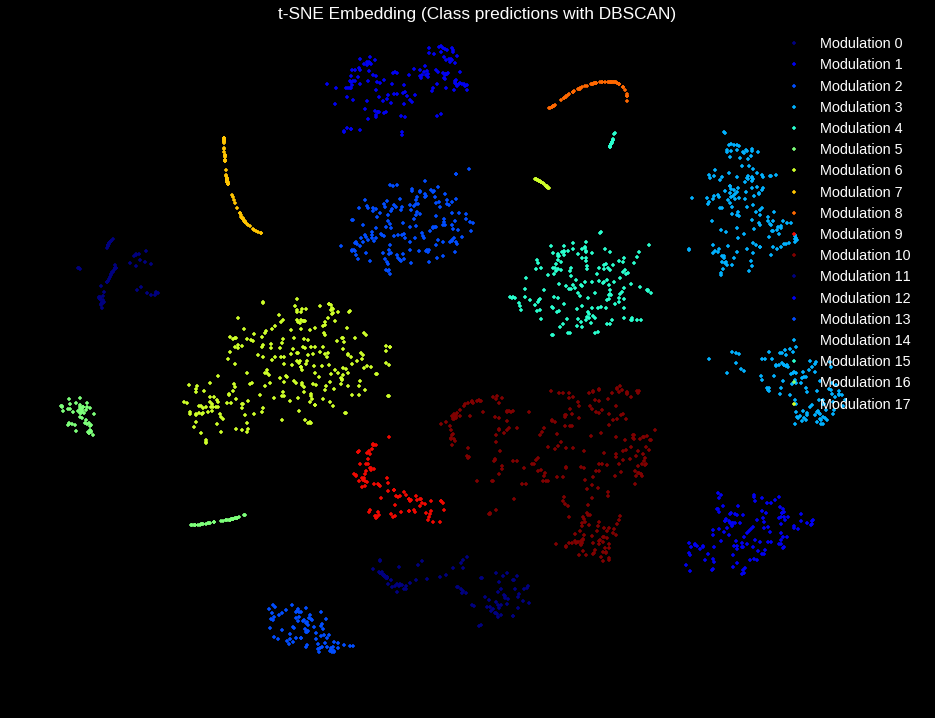}
    \caption{Blind Discovery of Modulation Classes using DBSCAN}
    \label{figure:ss6}
\end{figure}

Once examples form relatively separable clusters in the embedding space we may use any number of clustering algorithms to group and assign each of them to a class label.  In figure \ref{figure:ss6} we show an example using the DBSCAN \cite{dbscan} clustering algorithm to group clusters into a set of unknown but distinct modulation classes.  We find this clustering method to be relatively well suited to the obscurely shaped clusters formed in our compressed space.  Data curation after clustering may then consist of labeling clusters of many examples rather than each individual example, providing orders of magnitude in efficiency improvement from human data curation time.  On very large datasets with many classes and examples this makes managing large scale learning tasks much more tractable than they might otherwise be.

While these clusters are not error free, we can generally find a one to one or several to one mappings from discovered class clusters to distinct real named classes with relatively low error rate in this example.  This holds promise that such a method could be used in the future to quickly organize and label large sets of radio emissions, and to leverage prior knowledge about emitter class features, but still allow scaling of recognition system capabilities, features and class labels over time while minimizing the manual human labor required to do so. 

\section{Conclusion}

In this work, we have demonstrated that low level time series features learned with convolutional neural networks on raw sampled radio time-series data can be used to effectively cluster numerous radio signal modulation types without explicitly labeled training data.  We have shown that by leveraging compressed representations learned from both  discriminative feature maps and compressed reconstruction spaces, we can begin to organize and structure complex radio signals datasets with unlabeled or poorly labeled starting points.  This is a powerful result in that it demonstrates a potential way forward for learning to differentiate, reason, recall, and describe new and unknown radio signals without the need for manual curation or expert guidance.  This is a key requirement as we seek to build systems which scale and grow in capability from experience over time.  
Generalization of the compressed feature space bases to new signal types remains a key challenge in this domain, but we have shown in this work that this feature generalization does occur in some cases.  Going forward, attempts to quantify and optimize this effect will be important.

\section{Future Work}

There is much left to learn about how to effectively build semi-supervised radio signal recognition systems, but we hope this work has helped demonstrate several promising new ways forward in the field.  
There are a number of important areas for improvement of these methods including investigations of how best to learn discriminative bootstrap features which generalize well to unknown modulation types, methods to iteratively re-train supervised features on separable clusters, and other methods for optimizing feature space representations and clustering methods for best class separability.
While much of this work has been a qualitative exploratory effort, future work requires more quantitative performance measures and improved baseline measures such as with traditional forms of dimensionality reduction such as PCA, ICA, etc.  Such metrics could consist of correct class number estimation, full class confusion matrices and classification accuracy measures, and a number of other measures which have yet to be fully identified.
As we build towards a new class of radio sensing systems which can identify known radio classes and improve capabilities on-line to scale to new classes and new variations, these measures will be greatly important in optimizing such systems and scaling radio sense making to large datasets.

\appendices

\section*{Acknowledgment}

This work was supported in part by the Defense Advanced Research Agency under grant no. HR0011-16-1-0002.

\nocite{masci2011stacked}

\printbibliography

\end{document}